\documentclass[conference]{IEEEtran}
\IEEEoverridecommandlockouts
\usepackage[top=0.75in, bottom=1in, left=0.625in, right=0.625in]{geometry}

\usepackage[table,xcdraw]{xcolor}  % 放这行！
\usepackage{soul,framed}           % 不要再在这里重复 xcolor

\colorlet{shadecolor}{yellow}
\usepackage[pdftex]{graphicx}
\graphicspath{{../pdf/}{../jpeg/}}
\DeclareGraphicsExtensions{.pdf,.jpeg,.png}

\usepackage[cmex10]{amsmath}
\usepackage{amssymb}
\usepackage{array}
\usepackage{url}
\usepackage{amsfonts}
\usepackage[mathscr]{euscript} % 替换默认 mathscr 样式

\usepackage{multirow}
\usepackage{caption}
\usepackage{subcaption}
\usepackage{cuted}
\usepackage{hyperref}
\usepackage{cleveref}
\usepackage{float}
\usepackage{setspace}
\usepackage{lipsum}

\begin{document}
% \setstretch{0.92}
% \bstctlcite{IEEEexample:BSTcontrol}

\title{SoraNav: Adaptive UAV Task-Centric Navigation via Zero-shot VLM Reasoning}

\author{Hongyu Song, Rishabh Dev Yadav, Cheng Guo, and Wei Pan
    \thanks{The authors are with the Department of Computer Science, The University of Manchester, United Kingdom 
    (emails: \{hongyu.song-3, rishabh.yadav, cheng.guo-5\}@postgrad.manchester.ac.uk; wei.pan@manchester.ac.uk).}
}
% \author{Anonymous Authors}
% The paper headers
% \markboth{}{Song \MakeLowercase{\textit{et al.}}: SoraNav: Adaptive UAV Task-Centric Navigation via Zero-shot VLM Reasoning}

\maketitle
% === ABSTRACT ====================================================================
% =================================================================================

\begin{abstract}
% Interpreting visual observations and natural language instructions for complex task execution remains a key challenge in robotics. Despite recent advances, executing language-driven navigation remains difficult, particularly for UAVs in small-scale 3D environments. Existing Vision-Language Navigation (VLN) approaches are predominantly designed for ground robots and cannot generalize to aerial tasks that require full 3D spatial reasoning. The emergence of large Vision-Language Models (VLMs) enables zero-shot semantic reasoning from visual and textual inputs, but these models lack spatial grounding and are not directly applicable to navigation. To address these limitations, \textbf{SoraNav} is introduced, an adaptive Unmanned Aerial Vehicle (UAV) navigation framework that integrates zero-shot (no task-specific training) VLM reasoning with geometry-aware decision making.  
% 1) Geometric priors are incorporated into image annotations to constrain the VLM action space and enhance decision quality;  
% 2) A hybrid switching strategy leverages navigation history to alternate between VLM reasoning and geometry-based exploration, mitigating dead-ends and redundant revisits;  
% 3) A PX4-based hardware–software platform, comprising both a digital twin and physical micro-UAV, enables reproducible evaluation.  
% Experimental results show that in 2.5D scenarios, our method improves Success Rate (SR) by 25.7\% and Success weighted by inverse Path Length (SPL) by 14\%. In 3D scenarios, it improves SR by 29.5\% and SPL by 18.5\% relative to the baseline.

Autonomous navigation under natural language instructions represents a crucial step toward embodied intelligence, enabling complex task execution in environments ranging from industrial facilities to domestic spaces. 
However, language-driven 3D navigation for Unmanned Aerial Vehicles (UAVs) requires precise spatial reasoning, a capability inherently lacking in current zero-shot Vision-Language Models (VLMs) which often generate ambiguous outputs and cannot guarantee geometric feasibility.
Furthermore, existing Vision-Language Navigation (VLN) methods are predominantly tailored for 2.5D ground robots, rendering them unable to generalize to the unconstrained 3D spatial reasoning required for aerial tasks in small-scale, cluttered environments. 
In this paper, we present SoraNav, a novel framework enabling zero-shot VLM reasoning for UAV task-centric navigation. To address the spatial-semantic gap, we introduce Multi-modal Visual Annotation (MVA), which encodes 3D geometric priors directly into the VLM's 2D visual input. To mitigate hallucinated or infeasible commands, we propose an Adaptive Decision Making (ADM) strategy that validates VLM proposals against exploration history, seamlessly switching to geometry-based exploration to avoid dead-ends and redundant revisits.
Deployed on a custom PX4-based micro-UAV, SoraNav demonstrates robust real-world performance. Quantitative results show our approach significantly outperforms state-of-the-art baselines, increasing Success Rate (SR) by 25.7\% and navigation efficiency (SPL) by 17.3\% in 2.5D scenarios, and achieving improvements of 39.3\% (SR) and 24.7\% (SPL) in complex 3D scenarios.

\end{abstract}

\begin{IEEEkeywords}
Machine Learning for Robot Control; AI-Based Methods; Deep Learning for Visual Perception
\end{IEEEkeywords}

\section{Introduction}
\label{sec:Introduction}

Autonomous navigation guided by natural language instructions represents a key capability for embodied intelligence, enabling complex task execution in environments ranging from industrial facilities to domestic spaces. In navigation tasks, this requires reasoning over 3D environments \cite{LI2026103624} while aligning actions with high-level task descriptions \cite{cui2025generatingvisionlanguagenavigationinstructions}, with applications spanning industrial inspection \cite{ren2024embodied}, assistive navigation \cite{balatti2023robotassistednavigationvisuallyimpaired}, and search-and-rescue \cite{song2022multi}. The problem is especially challenging for UAVs operating in small-scale 3D environments, where tight endurance constraints preclude the exhaustive exploration typical of classical methods.

Existing approaches—classical mapping~\cite{zhou2020fuelfastuavexploration,zhang2025falconfastautonomousaerial}, end-to-end learning~\cite{maksymets2021thda,ramrakhya2022habitat}, and modular pipelines~\cite{sathyamoorthy2024convoicontextawarenavigationusing,yokoyama2023vlfmvisionlanguagefrontiermaps}—each lack either language grounding, generalization, or scalable reasoning, limiting instruction-driven 3D navigation (cf.\ \Cref{sec:Related Works}).

%段三：应用层面 VLN 多集中在地面机器人，空中无人机 (UAV) 的研究不足。
% Existing VLN research has predominantly focused on ground robots \cite{maksymets2021thda}, \cite{ramrakhya2022habitat} , which inherently constrains the planned trajectories to 2.5D paths that cannot fully exploit the 3D structure of the environment. Consequently, methods developed in dataset-based simulators such as Habitat cannot be directly transferred to UAVs and are incapable of accomplishing 3D navigation, for instance, reaching the second floor of a building in the absence of stairs. In addition, prior UAV-based VLN studies have primarily targeted long-range navigation from high-altitude viewpoints \cite{wang2024realisticuavvisionlanguagenavigation}, which are more suitable for city-scale monitoring tasks rather than everyday industrial or domestic scenarios such as factory floors or small parks.
% Existing VLN research has predominantly focused on ground robots \cite{maksymets2021thda, ramrakhya2022habitat}, constraining trajectories to 2.5D and limiting exploitation of full 3D spatial structure. As a result, methods developed in simulators such as Habitat cannot generalize to UAVs, and fail in tasks like reaching elevated targets without stairs. Prior UAV-VLN studies \cite{wang2024realisticuavvisionlanguagenavigation} mainly address long-range navigation from high-altitude views, which suits city-scale monitoring but not small-scale industrial or domestic environments such as factory floors or small parks.
Furthermore, existing VLN research has yet to address small-scale UAV navigation. Most work targets ground robots \cite{ramrakhya2022habitat}, whose navigation policies do not generalize to aerial tasks requiring full 3D spatial reasoning. Prior UAV-VLN studies \cite{wang2024realisticuavvisionlanguagenavigation} mainly address long-range navigation from high-altitude views, which suits city-scale monitoring but not small-scale industrial or domestic environments such as factories or small parks.
\begin{figure}[t!]
    \centering
    \includegraphics[width=\linewidth,height=0.85\linewidth,keepaspectratio]{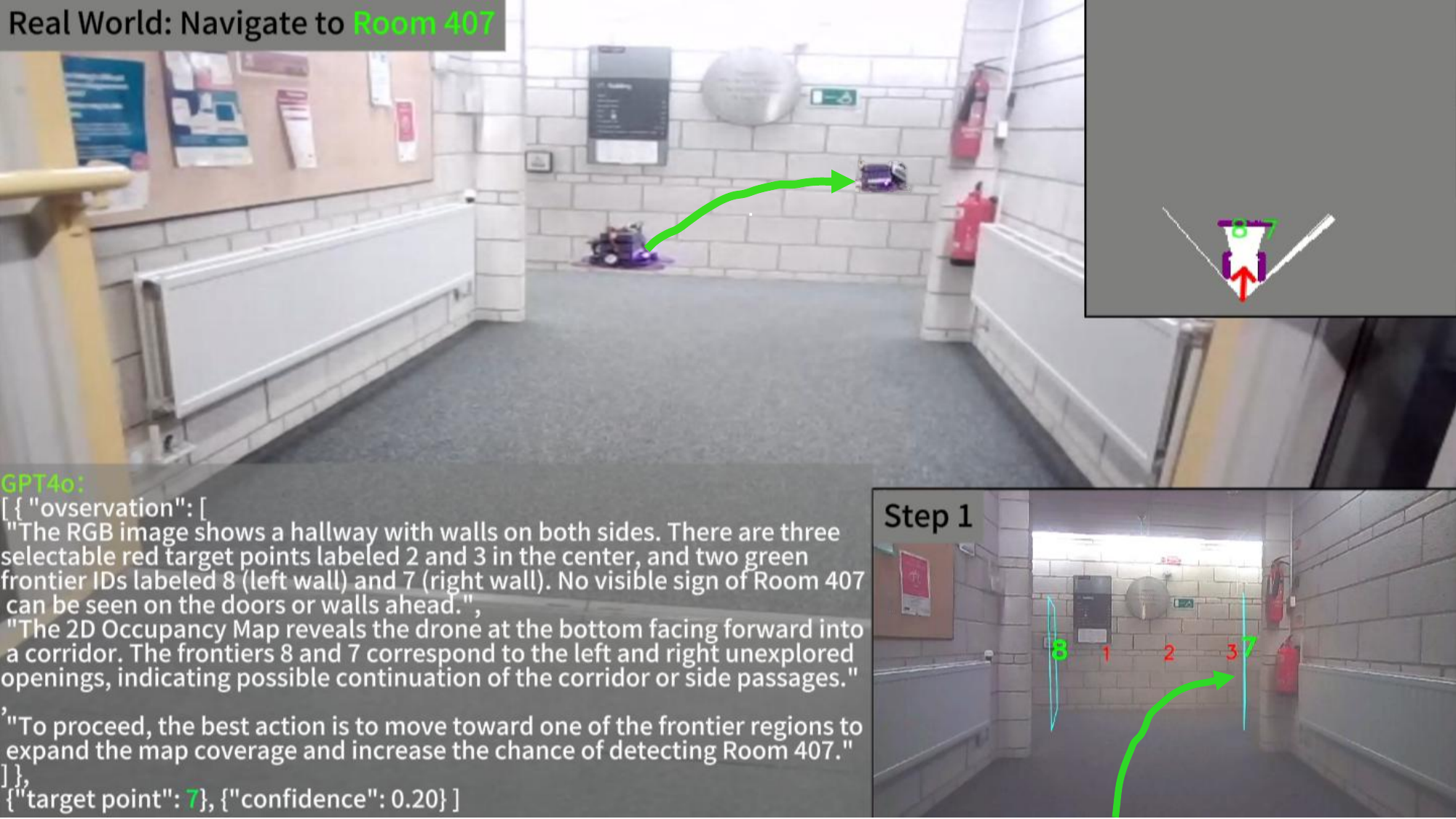}
    \caption{Depiction of a single-step navigation process. The VLM directs the UAV toward Frontier 7 to reveal more unknown areas.}
    \label{fig:cover}
\end{figure}

The recent emergence of large VLMs like GPT \cite{openai2023gpt4o} and Claude \cite{claude3} offers a promising alternative. Their powerful zero-shot semantic reasoning capabilities make it feasible to pursue \textbf{Zero-shot VLM-guided Task-Centric Navigation (ZSVTN)}: a navigation paradigm where a robot progressively explores and reaches a target described only by a natural language and visual instruction, without task-specific fine-tuning or reliance on rich semantic priors. Here, \emph{zero-shot} refers to the absence of any task-specific training or weight adaptation; a fixed, navigation-oriented prompt schema is used to structure the VLM query and output, but the model weights remain frozen throughout. However, general-purpose VLMs are not designed for navigation as they lack spatial grounding, generate ambiguous outputs, and cannot guarantee geometric feasibility. These limitations motivate a hybrid framework that couples VLM reasoning with geometry-aware decision-making for robust 3D UAV navigation in small-scale environments.
\begin{figure*}[!t]
    \centering
    \includegraphics[width=\linewidth]{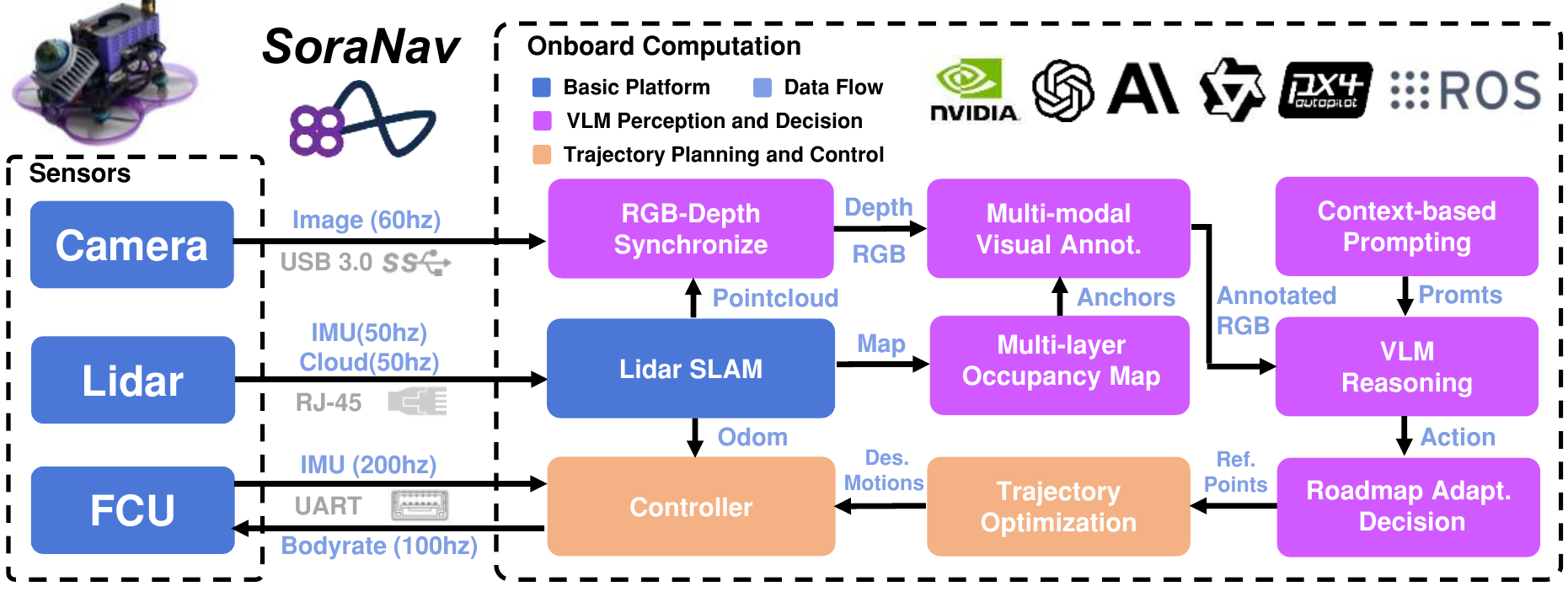}
    \caption{Illustration of the System Overview and Data Flow.}
    \label{fig:system_flow}
\vspace{-6mm}
\end{figure*}
%段五：引出本文的工作和贡献
% In this paper, we address these challenges by adopting an adaptive strategy similar to h1ow humans search for navigation objects. Human navigators typically approach targets or relevant landmarks once they appear in the field of view. When the target is not visible, they instead explore nearby unobserved regions while minimizing redundant visits to previously observed areas. Moreover, humans naturally select visual references as stopping points during navigation (e.g., stopping at the entrance of a classroom), rather than halting at a specific numerical coordinate. In short, such a navigation strategy resembles answering a multiple-choice question rather than an open-ended one. Building on these insights, we propose SoraNav, an adaptive small-scale UAV navigation framework using Zeroshot VLM reasoning and decision making, with 3 key contributions: 
% In this paper, we address these challenges by adopting an adaptive strategy inspired by how humans search for navigation targets. Human navigators typically approach targets or landmarks when they appear in view, and explore nearby unobserved regions when the target is not visible, while avoiding redundant revisits. They also rely on visual references as natural stopping points (e.g., pausing at a doorway) rather than predefined distances or coordinates. Building on these insights, we propose \textbf{SoraNav}, an adaptive small-scale UAV navigation framework that integrates zeroshot VLM reasoning with geometry-consistent decision making. Our contributions are threefold:

In this paper, we address these challenges by adopting an adaptive strategy inspired by human navigation behavior. Humans tend to move toward visible targets, explore nearby unknown areas when targets are not in sight, and avoid redundant revisits. Their decisions often rely on visual cues (e.g., doorways or intersections) as implicit waypoints, rather than explicit coordinates. Motivated by these insights, we present \textbf{SoraNav}, an adaptive navigation framework for small-scale UAVs that fuses zero-shot VLM reasoning with geometry-consistent decision making, as illustrated in \Cref{fig:cover} and \Cref{fig:system_flow}. Our contributions are summarized as follows:
\begin{itemize}
    \item \textbf{Multi-modal Visual Annotation (MVA):} 
    Geometric priors encoded as image annotations bridge VLMs' semantic strength with spatial grounding, reducing open-ended spatial inference to structured selection among three anchor types (target, frontier, and inter-layer).

    \item \textbf{Adaptive Decision Making (ADM):} 
    A hybrid switching strategy evaluates past navigation history to alternate between VLM reasoning and geometry-based exploration, avoiding dead-ends and redundant revisits.

    \item \textbf{Hardware–Software Platform:} 
    A PX4-based digital twin and a real micro-UAV setup are provided for ZSVTN, to be released as open source upon acceptance.
\end{itemize}

\section{Related Works}
\label{sec:Related Works}
%%逻辑梳理
%现有的自主导航方法，传统没语义，端到端过拟合难部署，混合方法reasoning不参与决策。 -> 上述弊端导致机器人无achieve instruction-driven 3D navigation from natural language task descriptions.
%现有的VLN研究集中于地面机器人无法直接在无人机上Sim2Real，之前的无人机VLN集中在大尺度环境。 -> 一个 Small-Scale UAV VLN框架被需要，并且需要有Sim2Real的能力。
%大VL模型有rich commonsense，可以分析场景，目标识别，并且具有reasoning能力。上述能力让ZS-VLSN变为可能，但是大VL模型无法直接输出可以被UAV直接执行的控制量，因此需要合理的映射。 大模型为了给出答案会产生幻觉。
% 插入图像：强制 figure 环境，但马上留白

%%正文
%%前三种方法弊端
% Classical, end-to-end, and modular approaches each offer distinct advantages in autonomous navigation. Classical approaches typically rely on a simultaneous localization and mapping (SLAM) module and active exploration to achieve complete navigation within a map.  
% For example, \cite{zhou2020fuelfastuavexploration} utilizes an incrementally updated frontier-based data structure for full scene exploration, while \cite{roh2025autonomous} integrates travel time and information gain to enable cross-floor navigation in unknown environments.  
% Some classical methods also incorporate semantic information; for instance, \cite{rosinol2020kimera} proposes a globally consistent 3D mapping framework with semantic labeling.  
% However, these approaches do not provide effective solutions for navigation tasks where goals are specified in semantic terms or natural language. 
Existing autonomous navigation methods fall into three categories. Classical methods \cite{zhou2020fuelfastuavexploration,roh2025autonomous,rosinol2020kimera} rely on SLAM and frontier-based exploration to build geometric maps but cannot interpret language-specified goals. End-to-end approaches \cite{maksymets2021thda,ramrakhya2022habitat,chen2023object} learn direct visual-to-action mappings with implicit semantic grounding, yet require extensive supervision and generalize poorly. Modular pipelines \cite{sathyamoorthy2024convoicontextawarenavigationusing,yokoyama2023vlfmvisionlanguagefrontiermaps,zhang2025apexnavadaptiveexplorationstrategy} integrate VLMs for language grounding with classical planners but still depend on multiple models, handcrafted strategies, and predefined labels. Collectively, these paradigms struggle to support instruction-driven 3D navigation in unstructured environments.

%UAV小尺度导航的空白。
% Existing VLN studies mainly address ground navigation, lacking solutions for small-scale UAVs and real-world transfer. 
% Ground robot-based works such as \cite{ramrakhya2022habitat} \cite{maksymets2021thda}, and \cite{yokoyama2023vlfmvisionlanguagefrontiermaps} are primarily developed within the Habitat simulation environment, which provides a well-supported ecosystem for reproducible research. 
% However, these approaches are difficult to migrate or deploy in aerial navigation scenarios due to the fundamentally different control dynamics and perception requirements of UAVs.  
% Recent UAV-VLN efforts \cite{wang2024realisticuavvisionlanguagenavigation}, \cite{gao2025openflycomprehensiveplatformaerial} have begun to explore the feasibility of language-guided UAV navigation and have introduced datasets built in high-fidelity simulation environments.  
% Nonetheless, these works focus on large-scale environments such as cities, where navigation tasks span long distances and success is defined with a relatively loose threshold (e.g., within 20 meters of the target).  
% Such benchmarks are unsuitable for small-scale scenarios like construction sites or indoor factories, where fine-grained control and close-range semantic understanding are critical.  
% This highlights the need for a high-fidelity small-scale UAV VLN framework capable of handling tasks such as inspection, search, and rescue in compact and cluttered environments.
Most VLN studies \cite{maksymets2021thda,ramrakhya2022habitat,yokoyama2023vlfmvisionlanguagefrontiermaps} assume ground-robot simulators (e.g., Habitat) with 2.5D motion and are difficult to transfer to aerial platforms due to differences in dynamics and perception.
Recent UAV-VLN efforts \cite{wang2024realisticuavvisionlanguagenavigation,gao2025openflycomprehensiveplatformaerial} target large-scale urban scenes with loose success thresholds (e.g., 20\,m), unsuitable for small-scale cluttered environments requiring precise short-horizon reasoning.
This gap motivates a UAV-VLN framework tailored to compact 3D spaces with instruction-driven navigation.

%引出本文对照文章
Large VLMs possess rich common sense knowledge, enabling them to perform scene understanding, object recognition, and even reasoning over navigation tasks, which makes ZSVTN feasible. In 2D navigation, \cite{pivot,navvlm} propose to annotate images with candidate targets and let the VLM select the most relevant one; experiments show that their annotated representations guide decision-making more effectively than raw images. 
However, these methods sample waypoints uniformly without geometric priors, lack validation of VLM decisions, and are restricted to 2D settings.
In 3D navigation, \cite{wang2024realisticuavvisionlanguagenavigation,gao2025openflycomprehensiveplatformaerial} utilize VLMs to infer spatial waypoints as navigation targets in real-world UAV scenarios, but due to hallucinations \cite{kalai2025languagemodelshallucinate} and a lack of explicit spatial scale awareness, VLMs can generate unsafe or unreasonable commands when reasoning from visual inputs alone. 
To address these gaps, SoraNav derives anchors from occupancy-based geometry with traversability guarantees instead of uniform sampling, assigns action semantics (approach, explore, switch layer) to each anchor, and validates VLM choices against exploration history via ADM.

\section{Problem Definition}
\label{sec:Algorithm}
We consider \textbf{ZSVTN} in unknown environments: a UAV progressively explores and reaches a target described only by natural language instructions, using a pre-trained VLM whose weights remain frozen.
We formulate this as a \textit{goal-conditioned sequential decision problem}.
The task is specified by a context-based prompt \(\mathcal{P}\) encoding the goal and behavioral constraints (\Cref{sec:adm}).
At each step \(k\), the agent maintains an information state
\begin{equation}
s_k \;=\; \bigl({}^G p_k,\; o_k,\; \mathcal{V}_{\mathrm{1:k-1}}^{\mathrm{rm}}\bigr),
\end{equation}
where \({}^G p_k\) is the drone pose in the global frame, \(o_k\) denotes the perceptual observation (e.g., RGB image, occupancy map), and \(\mathcal{V}_{\mathrm{1:k-1}}^{\mathrm{rm}}\) is the set of previously visited roadmap vertices summarizing the exploration history.
Given \(s_k\) and \(\mathcal{P}\), the high-level policy
\begin{equation}
a_k \;\sim\; \pi(\,\cdot \mid s_k,\, \mathcal{P}\,)
\end{equation}
selects a spatial anchor (Sections~\ref{sec:mva}--\ref{sec:adm}) mapped to a goal pose \({}^G p^{\mathrm{goal}}_k\).
This perception--decision--execution loop repeats until the target is visually confirmed.
The policy \(\pi\) is realised by the VLM via adaptive decision-making combining semantic reasoning with geometry-aware exploration (\Cref{sec:adm}), rather than explicit reward optimisation.

\section{Multi-modal Visual Annotation}
\label{sec:mva}
% - 深度+RGB对齐，滑窗法
% - 三种锚点的选取
%   - 地图表征（Z为其他分辨率）
%   - 同层collision-free target point
%   - 全局frontier视点
%   - 跨Height Layer  channel点
% - AR marking
% Since a VLM cannot directly infer geometric scales or frontier-related information from raw RGB images, we propose a \textbf{Multi-modal Visual Annotation (MVA)} scheme. 
Since a VLM cannot directly infer geometric scales or frontier-related information from raw RGB images, we propose a \textbf{Multi-modal Visual Annotation (MVA)} scheme to encode such information.
MVA derives anchors from real-time occupancy maps with traversability guarantees rather than uniform sampling.
Each anchor carries an action semantic (approach, explore, or switch layer), so the VLM selects both an action type and a goal.
Every decision is further validated against exploration history via ADM to filter revisits.
The observation is
\(o_{k}^{\mathrm{mva}} = (i_k^{\mathrm{mva}}, m_k^h)\), 
where \(i_k^{\mathrm{mva}}\) is the original RGB frame \(i_k\) annotated by spatial guiding anchors, 
and \(m^{h}_{k}\) is the 2D frontier map corresponding to the current flight altitude layer $h$. These annotations reveal traversability, unknown-space layout, and vertical navigability directly from a first-person view. An illustration of the anchors and visual annotation is shown in \Cref{fig:anchors_visual_annotations}.
\subsection{Depth Alignment}
To ensure that the anchors selected by the VLM from the RGB image correspond to precise spatial locations, we align LiDAR point clouds with RGB frames via extrinsic–intrinsic projection, thereby combining geometric accuracy with semantic perception.
Let ${}^B T_{\text{C}}$ and ${}^B T_{\text{lidar}}$ denote the camera and LiDAR poses 
in the body frame $B$, and ${}^G T_B$ the body pose in the global frame $G$. 
A LiDAR point ${}^G p_{\text{lidar}}$ is transformed into the camera frame $C$ as
\begin{equation}
{}^C p_{\text{lidar}} = ({}^B T_{\text{C}})^{-1} ({}^G T_B)^{-1} {}^G p_{\text{lidar}}.
\end{equation}
The resulting point ${}^C p_{\text{lidar}}=({}^C\!X, {}^C\!Y, {}^C\!Z)$ is projected onto the image plane via the pinhole model~\cite{newcombe2011kinectfusion}: $u = f_x {}^C\!X / {}^C\!Z + c_x$, $v = f_y {}^C\!Y / {}^C\!Z + c_y$, with depth $D(u,v) = {}^C\!Z$.
A sliding-window integration accumulates LiDAR point clouds to maintain a real-time aligned depth map.

\subsection{Multi-layer Occupancy Map}
To reduce annotation clutter and enable the VLM to interpret traversable regions across different heights, we encode the 3D scene as a multi-layer 2D occupancy map $\mathcal{M}=\{m^h\}$ \cite{s22103690}, where each $m^h$ is the 2D occupancy map at height layer $h$. Each cell is labeled as free, occupied, or unknown via raycasting. Frontiers are incrementally updated following \cite{zhou2020fuelfastuavexploration}, and for simplicity only the main plane and viewpoint of each frontier in the 2D height layers are considered.

% - 三种锚点的选取
\subsection{Spatial Guiding Anchors}
\subsubsection{Frontier Anchors}

\begin{figure}[t]
    \centering
    \includegraphics[width=\linewidth]{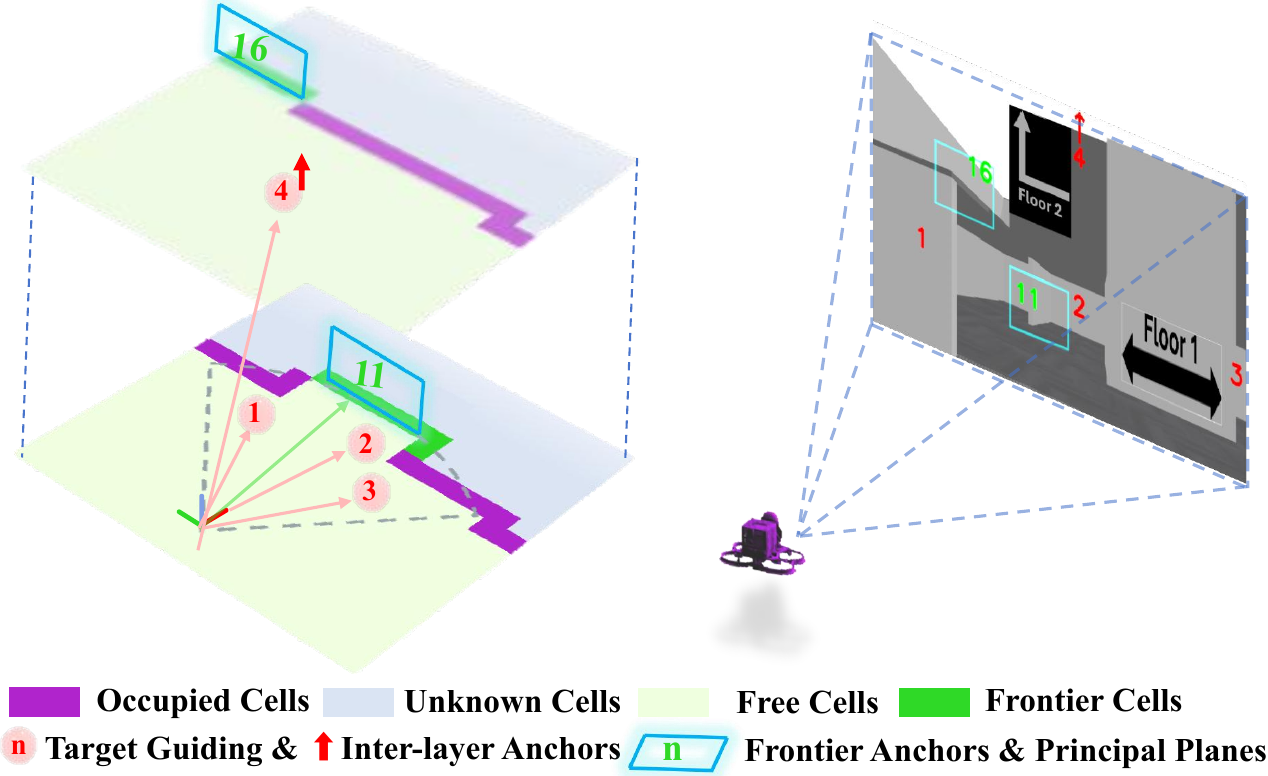}
    \caption{Illustration of Anchors and Visual Annotations.}
    \label{fig:anchors_visual_annotations}

\end{figure}
% as
% \begin{equation}
% {}^C \mathbf{w}_i = ({}^B T_{\text{cam}})^{-1} ({}^G T_B)^{-1}\mathbf{w}_i,
% \end{equation}
% \subset \mathscr{F}$
%要把frontier 投影到二维平面
The height of each frontier plane is determined by the spacing between adjacent height layers. 
For a frontier cluster $\mathscr{F}_c$  in a 2D occupancy map, 
we compute its principal axis $\mathbf{d}_c$ and define the planar length as $\ell_c$,
% \begin{equation}
% \ell_c = \max_{\mathbf{p},\mathbf{q}\in \mathscr{F}_c} 
% \langle \mathbf{p}-\mathbf{q}, \mathbf{d}_c \rangle .
% \end{equation}
if $\ell_c > \ell_{\max}$, the cluster is partitioned into smaller planar segments, where $\ell_{\max}$ denotes the maximum allowed length for a frontier cluster.
Each frontier plane is approximated by four corner points 
$\mathbf{w}_i = ({}^{G}x_i, {}^{G}y_i, {}^{G}z_i),\; i=1,\dots,4$ in the global frame.  
They are transformed into the camera frame and projected onto the image plane. Specifically, the FOV $\mathcal{F}({}^G p_k)$ is 
the spherical sector centered at ${}^G p_k$ with radius $d_{\max}$ and half-angle $\theta_{\max}$. 
If all projected corner points are inside the camera field of view, 
the corresponding frontier plane, defined by connecting its four corner points, is displayed on the RGB frame as an annotation. 
The contours of the frontier planes are shown in cyan, and their indices are labeled in green, as illustrated in \Cref{fig:anchors_visual_annotations}.
% {\color{red} ${}^{G}X_i, {}^{G}Y_i, {}^{G}Z_i$ ?? Not defined}

\subsubsection{Target Anchors}
Let $\mathscr{F}_{\mathrm{obs}}({}^G p_k) \subset \mathscr{F}({}^G p_k)$ denote the observable frontiers within $\mathcal{F}({}^G p_k)$, with count $n_{\mathrm{obs}}({}^G p_k)$. Given an angular span $[\varphi_{\mathrm{L}}, \varphi_{\mathrm{R}}]$, the \emph{adaptive sampling cardinality} is:
% \begin{equation}
% m_{\varphi} = 
% \begin{cases}
% m_{\max}, & n_{\mathrm{obs}}({}^G p_k) < \tau_{\mathrm{obs}}, \\
% m_{\mathrm{nom}}, & \text{otherwise},
% \end{cases}
% \label{eq:adaptive_cardinality}
% \end{equation}
\begin{equation}
m_{\varphi} = 
\left\{
\begin{aligned}
m_{\max}, & \quad \text{if } n_{\mathrm{obs}}({}^G p_k) < \tau_{\mathrm{obs}}, \\
m_{\mathrm{nom}}, & \quad \text{otherwise}.
\end{aligned}
\right.
\label{eq:adaptive_cardinality}
\end{equation}

where $\tau_{\mathrm{obs}}$ is the scarcity threshold, $m_{\mathrm{nom}}$ and $m_{\max}$ denote the nominal and maximum numbers of target directions. The discrete yaw set for target generation is then obtained via uniform discretization:
\begin{equation}
% \small
\Phi({}^G p_k) \triangleq
\left\{
\varphi_{\mathrm{L}} - \frac{l-1}{m_{\varphi}-1}(\varphi_{\mathrm{L}}-\varphi_{\mathrm{R}})
\;\middle|\; l\in\{1,\dots,m_{\varphi}\}
\right\},
\label{eq:yaw_sampling_set}
\end{equation}
% \begin{equation}
% % \Phi({}^G p_k)\triangleq\{\varphi_{\mathrm{L}}-\tfrac{l-1}{m_\varphi-1}(\varphi_{\mathrm{L}}-\varphi_{\mathrm{R}})\mid l=1{:}m_\varphi\}.
% % \label{eq:yaw_sampling_set}
% % \end{equation}
% \begin{equation}
% \small
% \Phi({}^G p_k)\triangleq\{\varphi_{\mathrm{L}}-\tfrac{l-1}{m_\varphi-1}(\varphi_{\mathrm{L}}-\varphi_{\mathrm{R}})\mid l=1{:}m_\varphi\}.
% \label{eq:yaw_sampling_set}
% \end{equation}

For each $\varphi \in \Phi({}^G p_k)$, we cast a ray from the sensor origin in $m^h_k$ and select the farthest LOS-reachable point with a safety clearance from occupied or unknown cells.

% This adaptive mechanism ensures that in observation-sparse regions ($n_{\mathrm{obs}}(v_j) < \tau_{\mathrm{obs}}$), the sampling resolution in yaw space is increased, thereby expanding exploratory coverage while maintaining collision-free guarantees.

\subsubsection{Inter-Layer Anchors}
During 3D navigation, the blind zones above and below the UAV increase the collision risk during altitude adjustments. To mitigate this, we introduce Inter-Layer Anchors, which are activated when collision-free navigation paths exist between adjacent height layers (e.g., they remain inactive if a ceiling is present above the UAV).

\section{Adaptive Decision Making \& Trajectory Planning}
\label{sec:adm}
\begin{figure*}[t]
    % \centering
    % 替换为你的图片路径
    \includegraphics[width=\linewidth]{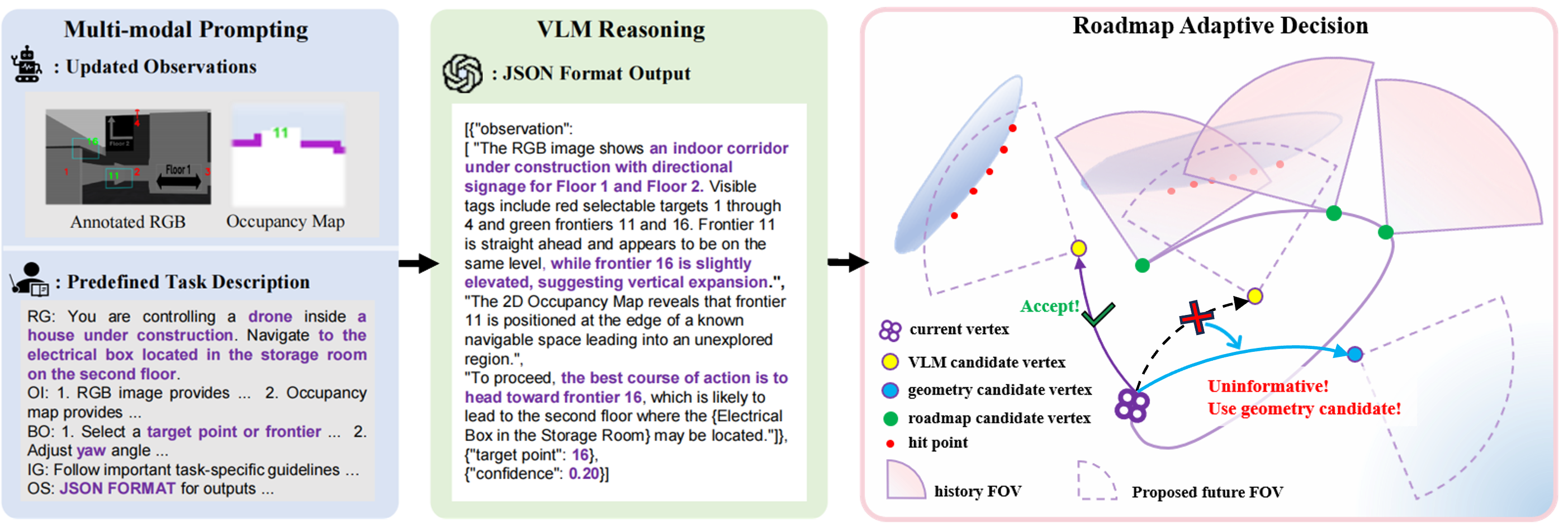}
    \caption{
        \textbf{Pipeline of Adaptive Decision Making.}
        Multi-modal prompting guides the VLM in reasoning and decision generation.
        A roadmap hypergraph validates VLM decisions, enabling a transition from \textit{uninformative} decisions to \textit{geometric} strategies.
    }
    \label{fig:adaptive_decision_pipeline}
\vspace{-2mm}
\end{figure*}

To enable effective navigation under VLM reasoning, we propose an \textbf{Adaptive Decision Making (ADM)} mechanism. 
Based on the MVA module, multi-modal prompting is first performed to obtain semantic proposals from the VLM. 
The prompting results are then validated by the Roadmap Adaptive Decision process, which determines whether to follow the VLM’s suggestion or navigate toward nearby unexplored regions for re-prompting. 
The final decision is converted into waypoints for trajectory generation and subsequent control commands. 
An overview of the ADM pipeline is illustrated in \Cref{fig:adaptive_decision_pipeline}.

\subsection{Multi-modal Prompting}
\subsubsection{Context-Based Prompt}
Inspired by \cite{amatriain2024promptdesignengineeringintroduction}, we design a structured context-based prompt $\mathcal{P} = \{\emph{RG}, \emph{OI}, \emph{IG}, \emph{BO}, \emph{OS}\}$ for querying the VLM. 
Specifically, \emph{RG} (Role and Goal) defines the operational role and mission objective; 
\emph{OI} (Observation Input) specifies the components of $o_{k}^{\mathrm{mva}}$ and instructs the VLM on their interpretation; 
\emph{IG} (Important Guidelines) encodes task-specific constraints for prioritizing MVA; 
\emph{BO} (Behavior Options) defines permissible actions, i.e., selecting a target $g_k$ or a yaw adjustment $\Delta\psi_k$; 
and \emph{OS} (Output Schema) specifies the structured output format for downstream processing.

\subsubsection{Prompting the VLM}
The VLM is queried with observation $o_{k}^{\mathrm{mva}}$ and prompt $\mathcal{P}$:
% \begin{equation}
% (o_k^{\mathrm{intp}},\, a^{\mathrm{ref}}_k,\, c_{\mathrm{det}}) = 
% \mathrm{VLM} 
% \bigl(o_{k}^{\mathrm{mva}} \,\big|\,
% \mathcal{P}\bigr),
% \end{equation}
\begin{equation}
(o_k^{\mathrm{intp}},\, a^{\mathrm{ref}}_k,\, c_{\mathrm{det}}) = 
\mathrm{VLM}\left(
o_{k}^{\mathrm{mva}} \,\middle|\, \mathcal{P}
\right)
\end{equation}
where $o_k^{\mathrm{intp}}$ is the VLM's interpretation, $a^{\mathrm{ref}}_k$ the proposed action, and $c_{\mathrm{det}} \in (0,1)$ the detection confidence, clipped as $\tilde{c}_{\mathrm{det}} = \mathrm{clip}(c_{\mathrm{det}}, \varepsilon, 1 - \varepsilon)$ with $\varepsilon=10^{-2}$.

% the execution standards specified in

\subsection{Roadmap Adaptive Decision}
\subsubsection{Hypergraph Formulation}
We define the hypergraph space  
\(\mathcal{H} = (\mathcal{V}^{\mathrm{rm}}, \mathcal{E})\),  
where \(\mathcal{V}^{\mathrm{rm}}\) is the set of \textit{Roadmap Vertices} that encode full decision states,  
and \(\mathcal{E}\) is the set of decision-driven hyperedges.  
Each hyperedge \(e_k \in \mathcal{E}\) encapsulates the \(k\)-th decision step, linking the current Roadmap Vertex \(v^{\mathrm{rm}}_{k}\) to its candidates originating from different sources:
% \begin{equation}
% e_k = 
% \bigl\{ \,
%    v^{\mathrm{rm}}_{k-1}, \;
%    v^{\mathrm{rm}}_{k}, \;
%    \{ v^{\mathrm{vlm}}_{k},\, v^{\mathrm{geo}}_{k} \}
% \bigr\},
% \end{equation}
\begin{equation}
e_k = 
\left( \,
   v^{\mathrm{rm}}_{k-1}, \;
   v^{\mathrm{rm}}_{k}, \;
   \left\{ v^{\mathrm{vlm}}_{k},\, v^{\mathrm{geo}}_{k} \right\}
\right)
\end{equation}

Here \(v^{\mathrm{rm}}_{k}\) is the roadmap vertex at step \(k\),
\(v^{\mathrm{vlm}}_{k} \in \mathcal{C}^{\mathrm{vlm}}_{k}\) is the VLM-proposed candidate via MVA,
and \(v^{\mathrm{geo}}_{k} \in \mathcal{C}^{\mathrm{geo}}_{k}\) is the geometry-based candidate \cite{song2021view}.
Each vertex stores its global position and the derived yaw.

% To describe the executed trajectory between two Roadmap Vertices,  
% we introduce an auxiliary set of Intermediate Vertices \(\mathcal{V}^{\mathrm{int}}\).  
% These vertices do not belong to the main hypergraph vertex set \(\mathcal{V}^{\mathrm{rm}}\);  
% instead, they are parameterized samples attached to the hyperedge originating from a Roadmap Vertex \(v_p\).  

% For each \(v_p \in \mathcal{V}^{\mathrm{rm}}\),  
% we define an associated point set \(\mathcal{P}(v_p) \subset \mathcal{V}^{\mathrm{int}}\) as:
% \begin{equation}
%     \mathcal{P}(v_p) = \{ v^{\mathrm{int}}_1, v^{\mathrm{int}}_2, \ldots, v^{\mathrm{int}}_{M_p} \}, \qquad 
% v^{\mathrm{int}}_j = (p_j).
% \end{equation}

% Thus, the hypergraph topology is solely constructed from  
% \((\mathcal{V}^{\mathrm{rm}}, \mathcal{E})\),  
% while \(\mathcal{V}^{\mathrm{int}}\) serves as an auxiliary set that enriches the geometric representation of each hyperedge without introducing independent nodes.
\subsubsection{Candidate Validation}
For each step $k$
, the informativeness of a VLM-proposed candidate 
$v_k^{\mathrm{vlm}}$ is quantified by comparing its raycast 
observations with the historical FOV coverage.

% Formally, we denote it as
% \begin{equation}
% \mathcal{F}(v_j) \triangleq 
% B(p_j,d_{\max}) \,\cap\, C(p_j,\psi_j,\theta_{\max}),
% \label{eq:fov}
% \end{equation}
% where $B(p_j,d_{\max})$ is the sphere of radius $d_{\max}$ centered at $p_j$,  
% and $C(p_j,\psi_j,\theta_{\max})$ is the cone centered at $p_j$ along direction $\psi_j$ 
% with half-angle $\theta_{\max}$.

Instead of all historical vertices, we only consider the $K$ nearest visited 
Roadmap Vertices to $v_k^{\mathrm{vlm}}$, denoted by 
$\mathcal{N}_K(v_k^{\mathrm{vlm}}) \subset \mathcal{V}_{\mathrm{1:k-1}}^{\mathrm{rm}}$.
The local past coverage is
\begin{equation}
\mathcal{F}_{\mathrm{past}}(v_k^{\mathrm{vlm}}) \triangleq 
\bigcup_{v_j \in \mathcal{N}_K(v_k^{\mathrm{vlm}})} 
\mathcal{F}(v_j),
\label{eq:fov_past}
\end{equation}

% Let $\mathrm{Occ}:\mathbb{G} \to \{0,1\}$ be the occupancy indicator. 
% For each uniformly sampled \emph{azimuth angle} $\theta_m$ within $\mathcal{F}({}^G p_k)$, define the \emph{discrete ray index sequence}
% \[
% \mathscr{R}\!\left(v_i^{\mathrm{vlm}},\theta_m\right) \triangleq 
% \big\{ g_\ell \in \mathbb{G} \ \big|\ \ell\in\mathbb{N}^+ \big\}
% \]
% as the ordered set of grid cell centers generated by the \emph{Bresenham integer line algorithm}~\cite{bresenham1965} 
% from the viewpoint position $p_i$ along the direction 
% $\mathbf{d}_m = [\cos\theta_m,\ \sin\theta_m]^\top$.  
% The \emph{hit point} for the $m$-th ray is then defined as
% \begin{equation}
% h_i^m \triangleq \mathrm{center}\!\left(
% \underset{g_\ell\in\mathscr{R}(v_i^{\mathrm{vlm}},\theta_m)}{\arg\min} \ \ell \ \ \text{s.t.} \ \ \mathrm{Occ}(g_\ell)=1
% \right),
% \label{eq:hitpoint}
% \end{equation}
% where $\mathrm{center}(\cdot)$ maps a grid index to its Cartesian coordinates.  
% The set of all such first-hit points for $v_i^{\mathrm{vlm}}$ is denoted 
% $\mathcal{H}_i \triangleq \{h_i^1,\dots,h_i^{M_i}\}$.
% % 
% Let $\mathrm{Occ}: m^h \to \{-1,0,1\}$ be the occupancy indicator of the current layer map,
% where $\mathrm{Occ}(g)=-1$ indicates an \emph{unknown} cell,
% $\mathrm{Occ}(g)=0$ a \emph{free} cell,
% and $\mathrm{Occ}(g)=1$ an \emph{occupied} cell.
% For each azimuth angle $\theta_m$ sampled within $\mathcal{F}({}^G p_k)$,
% define the discrete ray
Let $\mathrm{Occ}: m^h \to \{-1,0,1\}$ denote the occupancy ($-1$: unknown, $0$: free, $1$: occupied).
For each azimuth $\theta_m$ sampled within $\mathcal{F}({}^G p_k)$, the discrete ray generated by Bresenham's algorithm~\cite{bresenham1965} from ${}^G p_k$ along $\mathbf{d}_m=[\cos\theta_m,\sin\theta_m]^\top$ is
\begin{equation}
\mathscr{R}\!\left(v_k^{\mathrm{vlm}},\theta_m\right) \triangleq 
\big\{ g_\ell \in \mathbb{G} \ \big|\ \ell\in\mathbb{N}^+ \big\},
\end{equation}
and the \emph{hit point} is the first non-free cell encountered:
\begin{equation}
h_k^m = \mathrm{center}\!\left(g_{\ell^\star}\right), \;
  \ell^\star = \min\!\left\{\, \ell \mid \mathrm{Occ}(g_\ell) \in \{-1,1\} \,\right\}.
\label{eq:hitpoint}
\end{equation}
The full hit-point set at step $k$ is $\mathcal{H}_k=\{h_k^1,\dots,h_k^{M_k}\}$.
% Each $h_i^m$ is assigned a coverage probability
% \begin{equation}
% P_{\mathrm{seen}}(h_i^m) \triangleq \max_{v_j \in \mathcal{N}_K(v_k^{\mathrm{vlm}})} 
% \mathbb{1}\!\left(h_i^m \in \mathcal{F}(v_j)\right),
% \label{eq:seen}
% \end{equation}
% Let $s_k^m \in \{0, 1\}$ denote the binary visibility indicator of $h_k^m$, defined as
% \begin{equation}
% s_k^m \triangleq 
% \max_{v_j \in \mathcal{N}_K(v_k^{\mathrm{vlm}})} 
% \mathbb{1}\!\left(h_k^m \in \mathcal{F}(v_j)\right),
% \label{eq:seen}
% \end{equation}
% where $\mathbb{1}(\cdot)$ is the indicator function.
% Then the information gain is then defined as
% \begin{equation}
% G(v_i^{\mathrm{vlm}}) \triangleq 
% \frac{1}{M_i} \sum_{h_i^m \in \mathcal{H}_i} \left(1 - s_i^m\right).
% \label{eq:gain}
% \end{equation}

% % --- New formula: normalize G into probability ---
% We normalize the information gain into a probabilistic form:
% \begin{equation}
% P_G(v_i^{\mathrm{vlm}}) \triangleq \sigma\big(\alpha\,(G(v_i^{\mathrm{vlm}}) - \tau_G)\big),
% \label{eq:pg}
% \end{equation}
% where $\alpha > 0$ controls the slope and $\tau_G$ shifts the geometric gain threshold.
Let $s_k^m \in \{0, 1\}$ denote the binary visibility indicator of $h_k^m$, defined as
\begin{equation}
s_k^m \triangleq 
\max_{v_j \in \mathcal{N}_K(v_k^{\mathrm{vlm}})} 
\mathbb{1}\!\left(h_k^m \in \mathcal{F}(v_j)\right),
\label{eq:seen}
\end{equation}
where $\mathbb{1}(\cdot)$ is the indicator function.
The information gain at $v_k^{\mathrm{vlm}}$ is
\begin{equation}
G(v_k^{\mathrm{vlm}}) \triangleq 
\frac{1}{M_k} \sum_{h_k^m \in \mathcal{H}_k} \left(1 - s_k^m\right),
\label{eq:gain}
\end{equation}
which measures the proportion of newly observed hit points.
We map the gain to a confidence score via a logistic function:
\begin{equation}
P_G(v_k^{\mathrm{vlm}}) \triangleq \sigma\big(\alpha\,(G(v_k^{\mathrm{vlm}}) - \tau_G)\big),
\label{eq:pg}
\end{equation}
where $\sigma(x)=(1+e^{-x})^{-1}$, $\mathrm{logit}(p)=\log\frac{p}{1-p}$, $\alpha > 0$ controls the slope, and $\tau_G$ is the gain threshold.
The validation probability is computed via log-odds fusion \cite{hornung2013octomap}:
\begin{equation}
P_{\mathrm{valid}}(v_k^{\mathrm{vlm}}) =
\sigma\!\left(\mathrm{logit}(P_G(v_k^{\mathrm{vlm}})) + \lambda \, \mathrm{logit}(\tilde{c}_{\mathrm{det}})\right),
\label{eq:valid}
    \end{equation}
where $\lambda > 0$ weights the detection confidence relative to the geometric gain, and both terms are fused additively in the log-odds domain to filter out uninformative candidates.
% $\sigma(x) \triangleq (1 + e^{-x})^{-1}$ is the logistic mapping converting 
% log-odds to probability, and
% \triangleq \log \frac{p}{1-p}
The target goal pose ${}^G p^{\mathrm{goal}}_k$ is computed from the selected anchor $a_k$ as
\begin{equation}
{}^G p^{\mathrm{goal}}_k = \mathscr{P}(a_k), \;
a_k =
\left\{
\begin{aligned}
v^{\mathrm{vlm}}_{k}, & \quad \text{if } P_{\mathrm{valid}}(v^{\mathrm{vlm}}_{k}) > \tau_{\mathrm{valid}}, \\
v^{\mathrm{geo}}_{k}, & \quad \text{otherwise}.
\end{aligned}
\right.
\label{eq:gk}
\end{equation}
% where \(\mathscr{P}(\cdot)\) denotes the mapping from the selected hypergraph node to its corresponding spatial pose, and \(\tau_{\mathrm{valid}}\) is the validity threshold. 
% The final yaw of ${{}^G p_{goal}}_k$ is determined based on detection confidence and directional deviation. 
% When $\tilde{c}_{\mathrm{det}} > \tau_{\mathrm{yaw}}$ and $|\psi_{\mathrm{traj}} - \psi_k| > \tau_{\Delta\psi}$, 
% the current yaw $\psi_k$ is retained to ensure target visibility; otherwise, 
% the yaw is aligned with the trajectory heading.    
where \(\mathscr{P}(\cdot)\) maps the selected hypergraph node to its spatial pose and \(\tau_{\mathrm{valid}}\) is the validity threshold. The geometric fallback is also triggered when $v^{\mathrm{vlm}}_{k}$ refers to a non-existent or infeasible anchor.
The final yaw of ${}^G p^{\mathrm{goal}}_k$ is set to $\psi_k$ when $\tilde{c}_{\mathrm{det}} > \tau_{\mathrm{yaw}}$ and $|\psi_{\mathrm{traj}} - \psi_k| > \tau_{\Delta\psi}$, and to the trajectory heading $\psi_{\mathrm{traj}}$ otherwise.

% 其中，\tau_{\mathrm{yaw}}$是保持保持yaw角的检测置信度阈值，\psi_{\mathrm{traj}}是轨迹末端的yaw角，\tau_{\Delta\psi}是判断yaw角变化的阈值
\subsection{Trajectory Generation}
To ensure that the decisions generated by the large VLM can be accurately executed, 
the UAV navigates toward the target goal pose ${}^G p^{\mathrm{goal}}_k$ considering both efficiency and safety. 
Given the current pose ${}^G p_k$, if ${}^G p^{\mathrm{goal}}_k$ originates from target anchors, 
a minimum-jerk trajectory is adopted, which is initialized by sampling along the straight line 
connecting ${}^G p^{\mathrm{goal}}_k$ and ${}^G p_k$. 
If it originates from frontier anchors, inter-layer anchors, or the geometry-based strategy, 
a trajectory generated following \cite{zhou2020ego} is employed instead.
The generated trajectory is then tracked by a controller through body-rate control.

\section{UAV Platform and Digital Twin}
\begin{figure}[t]
  \vspace{-7mm}
  \centering
  \begin{subfigure}[b]{\linewidth}
    \centering
    \includegraphics[width=\linewidth]{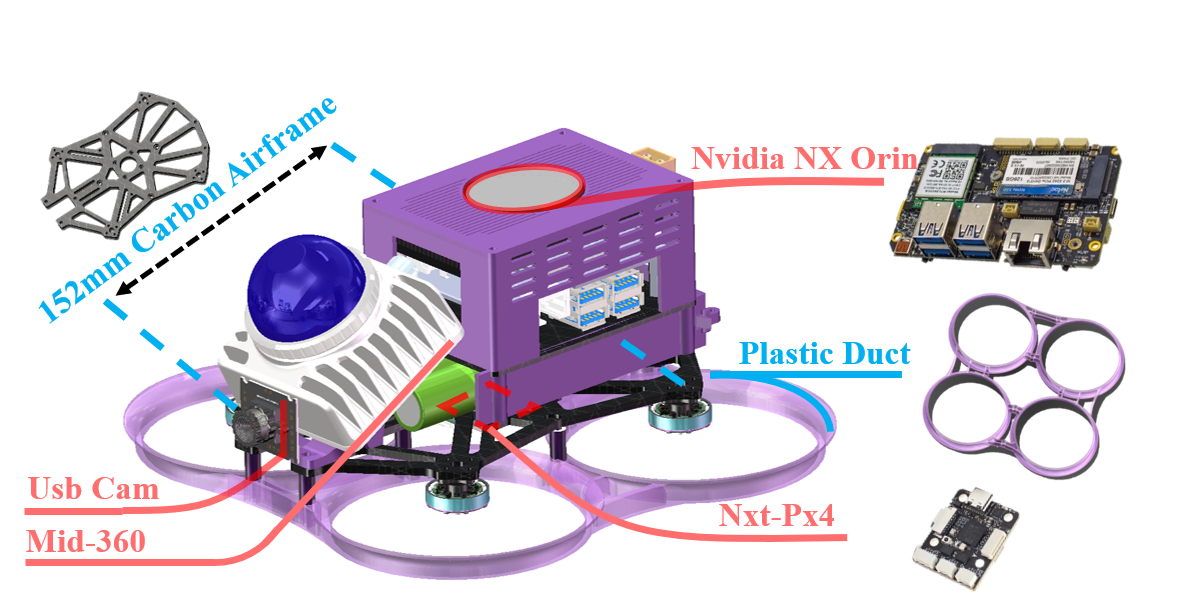}
    \caption{Custom UAV platform designed for ZSVTN.}
    \label{fig:uav_platform}
  \end{subfigure}
  \vspace{2mm}
  \begin{subfigure}[b]{\linewidth}
    \centering
    \includegraphics[width=\linewidth]{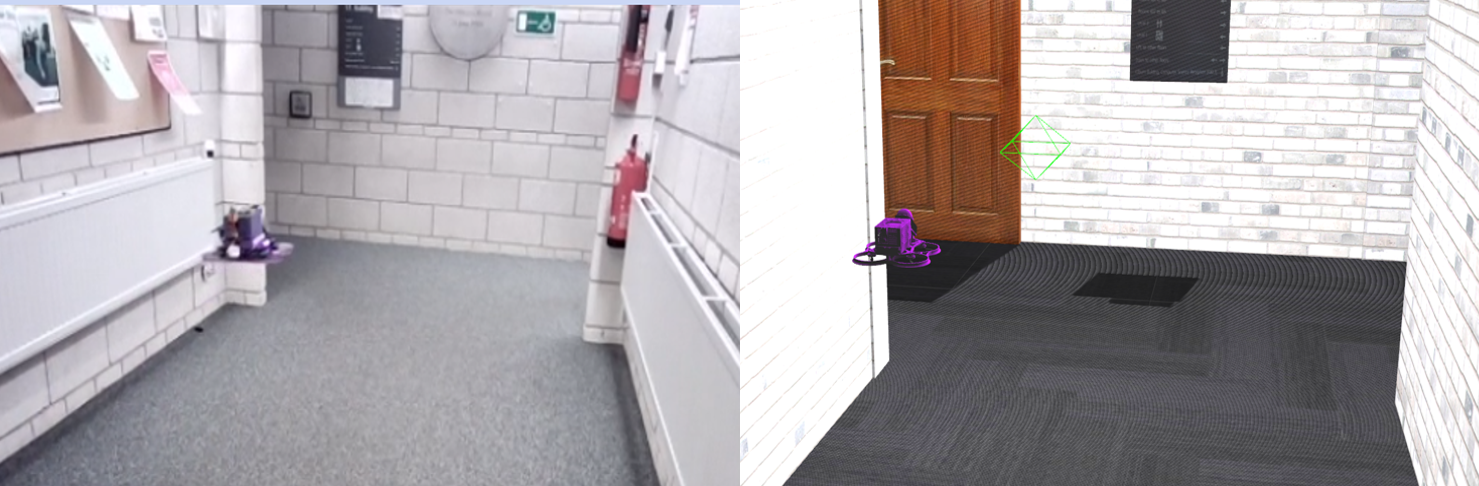}
    \caption{Real-world (left) and simulated (right) flight scenes.}
    \label{fig:real_vs_sim}
  \end{subfigure}
  \caption{UAV platform and digital twin environment.}
  \label{fig:platform_and_dt}
\end{figure}
A high-fidelity digital twin of the UAV is developed using PX4 SITL, ROS and livox-gazebo-plugin, enabling seamless transfer of the onboard software stack between simulation and the physical drone. \Cref{fig:real_vs_sim} illustrates an example of the matched real and simulated environments.

% A high-fidelity digital twin of the UAV is developed using ROS~\cite{quigley2009ros}, PX4 SITL~\cite{px4_sitl}, and  livox-gazebo-plugin~\cite{livox_laser_simulation}, enabling seamless transfer of the onboard software stack between simulation and the physical drone. \Cref{fig:real_vs_sim} illustrates an example of the matched real and simulated environments.

% NVIDIA Jetson 

To support the ZSVTN task in compact 3D environments, we designed a custom Micro Aerial Vehicle (MAV) inspired by the architecture of \cite{CUAstro2024}. The platform features a compact 152\,mm diagonal span to ensure agile obstacle-aware navigation in narrow spaces, shown in Figure~\ref{fig:uav_platform}. A Mid-360 LiDAR is mounted with a fixed tilt angle to enhance vertical field-of-view coverage, enabling robust real-time localization and mapping via \cite{xu2021fastlio2}, while preserving a low-profile form factor. An Orin NX is integrated to handle perception, mapping, and trajectory control onboard, while VLM inference is offloaded to the cloud.

% The airframe model, defined in SDF format, replicates the real UAV’s geometry and sensor placement, and is fully compatible with PX4 firmware. For visualization, we employ a 3D model of the actual platform. Sensor simulation is performed with hardware-matched specifications, including the livox-gazebo-plugin~\cite{livox_laser_simulation} to emulate the scanning pattern of the Mid-360 LiDAR. 

\section{Results and Evaluations}
\label{sec:Experiments}
\begin{table}[t!]
    \centering\small
    \renewcommand{\arraystretch}{0.9}
    \begin{tabular}{@{}clccc@{}}
    \hline
     & Method & SR $\uparrow$ & DtG $\downarrow$ & NRE ($\gamma\!=\!0.5$)$\downarrow$ \\ \hline
    \multirow{5}{*}{2.5D} & Ours & \textbf{0.83} & \textbf{1.62} & \textbf{0.65} \\
     & NavVLM & 0.67 & 2.62 & 0.74 \\
     & CONVOI & 0.33 & 3.38 & 0.83 \\
     & Spatial & 0.22 & 6.03 & 0.95 \\
     & Pivot & 0.00 & 6.69 & 1.00 \\ \hline
    \multirow{2}{*}{3D} & Ours & \textbf{0.67} & \textbf{2.80} & \textbf{0.84} \\
     & Spatial & 0.00 & 7.24 & 1.00 \\ \hline
    \end{tabular}
    \caption{Image Spatial Reasoning.}
    \label{tab:image_spatial_reasoning}
    \end{table}
\begin{table*}[t!]
\renewcommand{\arraystretch}{1.1}
\resizebox{\textwidth}{!}{%
\begin{tabular}{ccccccccccccccccc}
\hline
 &  & \multicolumn{4}{c}{Prompts $\downarrow$} & \multicolumn{4}{c}{Path Length $\downarrow$} & \multicolumn{4}{c}{DtG $\downarrow$}  & OR $\uparrow$ & SR $\uparrow$ & SPL$\uparrow$ \\ \cline{3-14}
Env \& Dims & Method & Avg & Std & Max & Min & Avg & Std & Max & Min & Avg & Std & Max & Min &  &  &  \\ \hline
 & Ours     & \textbf{1.14} & 0.35 & 2 & 1 & 8.10 & 1.31 & 11.02 & 6.72 & 1.93 & 0.35 & 2.30 & 1.11 & 0.86 & \textbf{0.86} & \textbf{0.67} \\
 & Spatial  & 2.86 & 0.35 & 3 & 2 & 4.78 & 2.51 & 7.30 & 0.02 & 4.62 & 2.13 & 8.74 & 2.41 & 1.00 & 0.29 & 0.24 \\
 & CONVOI   & 2.43 & 0.54 & 3 & 2 & 7.66 & 0.94 & 8.99 & 6.63 & 2.06 & 0.59 & 2.64 & 1.30 & 1.00 & 0.71 & 0.65 \\
 & NavVLM   & 2.43 & 0.54 & 3 & 2 & 9.95 & 0.04 & 10.00 & 9.91 & 1.42 & 0.03 & 1.44 & 1.39 & -- & -- & -- \\
\multirow{-5}{*}{Warehouse 2.5D} & Pivot & 2.57 & 0.54 & 3 & 2 & 6.95 & 2.05 & 9.48 & 3.75 & 3.53 & 1.84 & 6.86 & 1.56 & -- & -- & -- \\ \hline
 & Ours     & \textbf{1.57} & 0.98 & 3 & 1 & 9.23 & 3.01 & 15.78 & 7.32 & 2.87 & 0.53 & 3.67 & 2.35 & 1.00 & \textbf{0.71} & \textbf{0.39} \\
 & Spatial  & 2.86 & 0.35 & 3 & 2 & 7.94 & 1.16 & 9.90 & 6.04 & 4.35 & 1.31 & 6.18 & 2.63 & 1.00 & 0.14 & 0.07 \\
 & CONVOI   & 2.71 & 0.54 & 3 & 2 & 13.32 & 7.61 & 28.29 & 6.48 & 4.11 & 2.92 & 9.80 & 1.58 & 0.67 & 0.43 & 0.14 \\
 & NavVLM   & 3.00 & 0.00 & 3 & 3 & 6.32 & 2.05 & 10.58 & 5.13 & 5.76 & 0.41 & 6.02 & 4.90 & -- & -- & -- \\
\multirow{-5}{*}{Park 2.5D} & Pivot & 3.00 & 0.35 & 3 & 3 & 9.59 & 4.80 & 18.18 & 4.73 & 5.43 & 3.20 & 11.34 & 1.42 & -- & -- & -- \\ \hline
 & Ours     & \textbf{2.29} & 1.89 & 5 & 1 & 12.91 & 8.15 & 26.63 & 7.20 & 4.50 & 3.28 & 9.31 & 2.13 & 0.80 & 0.57 & 0.18 \\
\multirow{-2}{*}{Warehouse 3D} & Spatial & 5.00 & 0.00 & 5 & 5 & 7.36 & 2.10 & 10.94 & 3.77 & 5.96 & 2.71 & 8.34 & 3.05 & -- & -- & -- \\ \hline
 & Ours     & \textbf{2.14} & 2.07 & 5 & 1 & 14.13 & 6.14 & 24.67 & 9.36 & 3.59 & 4.32 & 11.85 & 1.27 & 1.00 & \textbf{0.71} & \textbf{0.56} \\
\multirow{-2}{*}{Park 3D} & Spatial & 4.14 & 0.98 & 5 & 3 & 11.78 & 2.06 & 14.35 & 9.50 & 2.75 & 0.78 & 3.65 & 1.45 & 0.75 & 0.43 & 0.31 \\ \hline
 & Ours     & \textbf{5.67} & 0.47 & 6 & 5 & 30.55 & 1.91 & 32.38 & 27.92 & 3.16 & 1.32 & 4.96 & 1.83 & 1.00 & \textbf{0.67} & \textbf{0.58} \\
 & Spatial  & 6.33 & 1.70 & 8 & 4 & 8.94 & 2.62 & 12.21 & 5.81 & 11.99 & 0.20 & 12.27 & 11.80 & -- & -- & -- \\
 & CONVOI   & 12.00 & 4.24 & 15 & 6 & 21.29 & 11.58 & 37.45 & 10.96 & 14.11 & 2.08 & 16.59 & 11.51 & -- & -- & -- \\
 & NavVLM   & 9.67 & 1.70 & 12 & 8 & 22.46 & 7.41 & 28.98 & 12.09 & 6.83 & 5.43 & 14.51 & 2.93 & 1.00 & 0.33 & 0.33 \\
\multirow{-5}{*}{Kilburn 2.5D} & Pivot & 15.00 & 0.00 & 15 & 15 & 22.40 & 11.22 & 37.99 & 12.09 & 12.41 & 6.40 & 19.00 & 3.74 & -- & -- & -- \\ \hline
 & Ours     & 6.00 & 0.82 & 7 & 5 & 32.46 & 6.31 & 41.38 & 27.90 & 4.75 & 2.42 & 8.11 & 2.50 & 1.00 & 0.33 & 0.31 \\
\multirow{-2}{*}{Construction 3D} & Spatial & 9.33 & 4.03 & 15 & 6 & 27.99 & 11.10 & 41.98 & 14.83 & 8.55 & 2.50 & 10.35 & 5.02 & -- & -- & -- \\ \hline
\end{tabular}%
}
\caption{Short-term and Long-term Navigation. Warehouse and Park correspond to short-term tasks, while the remaining scenarios represent long-term tasks. Note that we only bold the key metrics that effectively reflect task completion.}
\label{tab:navigation}
\end{table*}
\begin{table*}[t!]
\renewcommand{\arraystretch}{1.05}
\resizebox{\textwidth}{!}{%
\begin{tabular}{llccccccccccccccc}
\hline
 &  & \multicolumn{4}{c}{Prompts $\downarrow$} & \multicolumn{4}{c}{Path Length $\downarrow$} & \multicolumn{4}{c}{DtG $\downarrow$} & OR $\uparrow$ & SR $\uparrow$ & SPL $\uparrow$ \\ \cline{3-14}
Env \& Dims & Model & Avg & Std & Max & Min & Avg & Std & Max & Min & Avg & Std & Max & Min &  &  &  \\ \hline
\multirow{4}{*}{Kilburn 2.5D} 
 & GPT-4o    & \textbf{5.67} & 0.47 & 6 & 5 & 30.55 & 1.91 & 32.38 & 27.92 & \textbf{3.16} & 1.32 & 4.96 & 1.83 & 1.00 & \textbf{0.67} & \textbf{0.58} \\
 & Sonnet4   & 7.67 & 0.94 & 9 & 7 & 29.74 & 6.96 & 36.49 & 20.17 & 7.89 & 4.80 & 12.64 & 1.32 & 0.00 & -- & -- \\
 & Qwen2.5   & 6.00 & 4.24 & 9 & 0 & 16.59 & 5.59 & 23.94 & 10.39 & 14.78 & 3.17 & 18.97 & 11.30 & -- & -- & -- \\
 & Gemini2.5 & 6.33 & 0.94 & 7 & 5 & 30.59 & 4.09 & 35.29 & 25.32 & 7.36 & 4.20 & 11.03 & 1.47 & 1.00 & 0.33 & 0.21 \\ \hline
\multirow{4}{*}{Warehouse 3D} 
 & GPT-4o    & \textbf{2.29} & 1.89 & 5 & 1 & 12.91 & 8.15 & 26.63 & 7.20 & 4.50 & 3.28 & 9.31 & 2.13 & 0.80 & \textbf{0.57} & 0.18 \\
 & Sonnet4   & 3.71 & 1.58 & 5 & 1 & 18.19 & 6.31 & 30.06 & 10.02 & \textbf{3.19} & 2.10 & 6.55 & 1.15 & 0.75 & 0.43 & \textbf{0.32} \\
 & Qwen2.5   & 3.43 & 1.68 & 5 & 1 & 10.01 & 1.78 & 12.43 & 6.67 & 4.84 & 2.84 & 9.66 & 1.34 & 0.33 & 0.14 & 0.13 \\
 & Gemini2.5 & 5.00 & 0.00 & 5 & 5 & 15.29 & 4.08 & 20.92 & 7.33 & 6.61 & 2.61 & 10.15 & 1.90 & 1.00 & 0.14 & 0.09 \\ \hline
\end{tabular}%
}
\caption{Performance Across Different VLMs.}
\label{tab:vlm_performance}
\end{table*}
\begin{table}[t!]
\centering\small
\renewcommand{\arraystretch}{1.05}
\begin{tabular}{lcccc}
\hline
Method & Prompts $\downarrow$ & DtG $\downarrow$ & SR $\uparrow$ & SPL $\uparrow$ \\ \hline
Ours & 5.67 & 3.16 & 0.67 & 0.58 \\
Ours w/o ADM & 8.33 & 6.21 & 0.33 & 0.23 \\
Ours w/o MVA & 8.00 & 10.10 & -- & -- \\ \hline
\end{tabular}
\caption{Ablation study on the Kilburn 2.5D scenario.}
\label{tab:ablation_kilburn}
\end{table}
% To validate the effectiveness of the proposed framework, we conduct extensive experiments. The evaluations are organized along four complementary dimensions: 1) the effect of MVA on visual spatial reasoning, 2) the short-term and long-term navigation performance enabled by ADM, and 3) the generalizability across different large vision-language models. 4) a real-world deployment.
%We designed experiments to quantitatively evaluate the contribution of each component in SoraNav and to validate its scalability from simulation to real-world deployment.
%Specifically, the evaluations are organized along four complementary objectives:
We conducted experiments to assess each SoraNav component and validate its sim-to-real transfer, organized along four complementary objectives:
(1) how MVA enhances spatial reasoning from single-image interpretation;
(2) how ADM affects the short- and long-horizon navigation performance;
(3) generalizability across different VLMs; and
(4) overall system performance through real-world deployment on a real UAV.

\subsection{Experimental Setup}
% 环境 Evaluation Metrics: Implementation Details: Baselines:
\paragraph{Evaluation Scenarios} 
% We design the following scenarios for the subsequent experiments.  
% \textbf{Image Spatial Reasoning:} 6 x 2.5D scenes (6 outdoor, 2 indoor) and 4 x 3D scenes (3 outdoor, 1 indoor).  
% \textbf{Short-term \& Long-term Navigation Tasks:} 4 short-term scenarios (2 in 2.5D, 2 in 3D) and 3 long-term scenarios (1 in 2.5D, 2 in 3D).
We evaluate across four settings:
(1)~image spatial reasoning using 6 outdoor and 2 indoor 2.5D scenes plus 3 outdoor and 1 indoor 3D scenes;
(2)~navigation tasks comprising 4 short-horizon (2 in 2.5D, 2 in 3D) and 2 long-horizon (1 in 2.5D, 1 in 3D) scenarios;
(3)~model generalization across four VLMs (\texttt{GPT-4o}, \texttt{Sonnet4}, \texttt{Qwen2.5}, \texttt{Gemini2.5});
(4)~real-world transfer with deployment on a physical UAV.

\paragraph{Evaluation Metrics} 
% For Image Spatial Reasoning, we report \textbf{Success Rate} (SR), \textbf{Distance to Goal} (DtG) (m), and \textbf{Normalized Residual Error} (NRE), which is 
% \begin{equation}
% \mathrm{NRE} = \frac{1}{N} \sum_{i=1}^{N} \left( (1 - S_i) + S_i \cdot \left( \frac{D_i}{D_{\max}} \right)^{\gamma} \right).
% \end{equation}
% This metric captures both success and decision accuracy simultaneously, 
% where $S_i \in \{0,1\}$ indicates whether the decision succeeds, 
% $D_i$ denotes the spatial distance between the VLM-predicted position and the ideal target location, 
% and $D_{\max}$ is the maximum allowable error in successful cases.
For Image Spatial Reasoning, we report \textbf{Success Rate} (SR), \textbf{Distance to Goal} (DtG, m), and \textbf{Normalized Residual Error} (NRE), defined as 
$\mathrm{NRE} = \frac{1}{N} \sum_{i=1}^{N} \big( (1 - S_i) + S_i \cdot (D_i / D_{\max})^{\gamma} \big)$,
where $S_i \in \{0,1\}$ indicates whether the decision succeeds, $D_i$ denotes the spatial distance between the VLM-predicted position and the ideal target location, and $D_{\max}$ is the maximum allowable error in successful cases. This metric captures both success and decision accuracy simultaneously.

For navigation, we report SR, \textbf{Success weighted by inverse Path Length} (SPL), and DtG, where SPL evaluates efficiency relative to the optimal path.  
%For navigation tasks, 
Additionally, we record the number of VLM promptings and the path length (m) at episode termination.  
We also report \textbf{Observation Rate} (OR), i.e., whether the goal lies within the FOV upon first reaching the success distance.
A navigation episode is considered successful if the UAV reaches within 3\,m of the goal and the goal is visible.  
For short-term tasks, the maximum number of prompting steps is 3 in 2.5D and 5 in 3D, whereas long-term tasks allow up to 15 prompting steps.
When comparing our approach across different large models, 
long-term tasks are restricted to a maximum of 9 prompting steps.
An episode is also terminated early if a collision occurs or 3 consecutive commands are infeasible.

\begin{figure*}[t!]
    \centering
    \includegraphics[width=\textwidth]{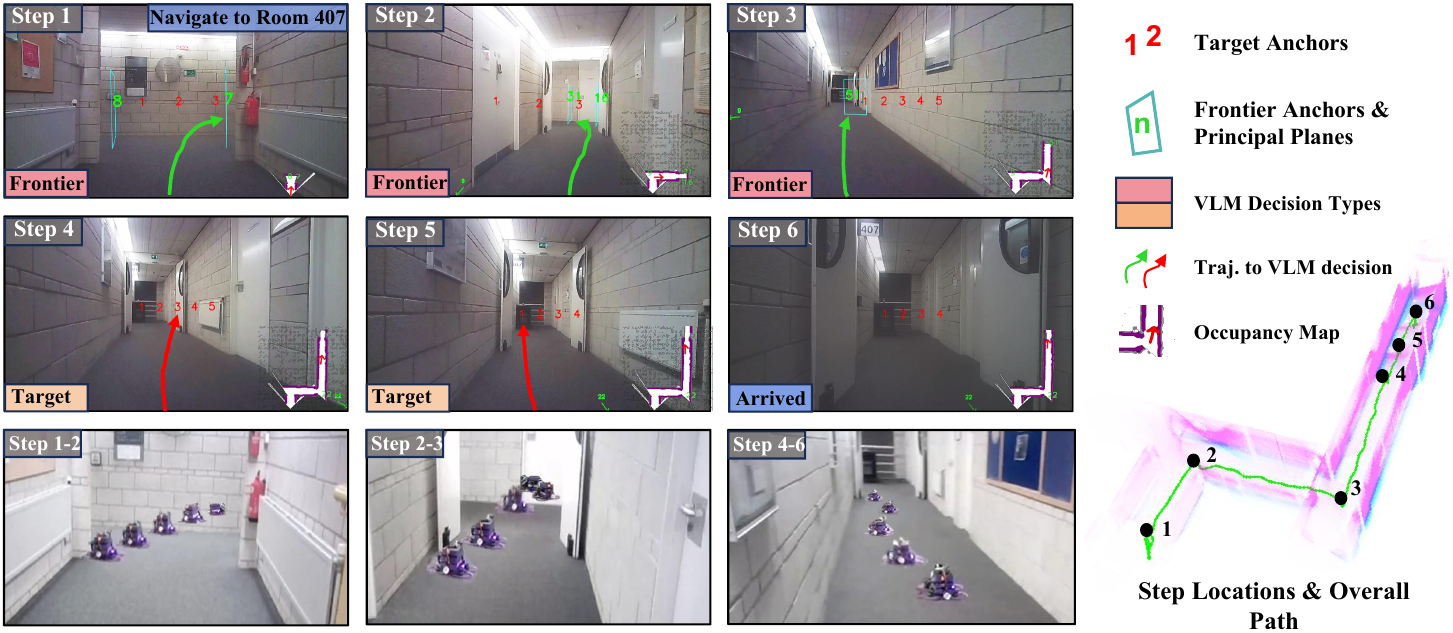}
    \caption{Real-world Demonstration. Autonomous navigation along an indoor corridor toward Room 407.}
    \label{fig:realworld_demo}
    \vspace{-6mm}
\end{figure*}
\paragraph{Baselines} 
%Our approach is compared with 
We consider the following four ZSVTN methods.  
\textbf{Spatial} follows \cite{gao2025openflycomprehensiveplatformaerial}, where the VLM directly de-tokenizes relative spatial positions, and thus serves as a baseline in both 2.5D and 3D environments.  
We further adapt three ground-robot navigation methods to aerial navigation: \textbf{NavVLM} \cite{navvlm}, \textbf{Pivot} \cite{pivot}, and \textbf{CONVOI} \cite{sathyamoorthy2024convoicontextawarenavigationusing}.  
Since these three methods do not account for altitude, they are evaluated only in 2.5D environments.  
NavVLM outputs collision-free points sampled at fixed angular intervals, while Pivot uniformly samples navigable waypoints; in both cases, sampled waypoints are annotated with indices on the image for the VLM to select from.  
CONVOI samples free cells in the occupancy grid map and requires the VLM to construct a path by sequentially connecting them.
All ADM parameters are fixed across every experiment: $\alpha = 8$, $\tau_G = 0.3$, $\lambda = 1.0$, $\tau_{\mathrm{valid}} = 0.6$, and $\tau_{\mathrm{yaw}} = 0.65$.

\subsection{Image Spatial Reasoning}

This experiment evaluates the impact of MVA on the spatial decision-making capability of vision-language models (VLMs). Each method receives the same RGB input and movement range, and must select a spatial anchor based on a natural language instruction containing both explicit and implicit spatial references (e.g., \textit{``between two trees"}, \textit{``near a shelf and a bin"}, \textit{``right-side window on the second floor"}).

\Cref{tab:image_spatial_reasoning} shows that MVA significantly improves anchor selection accuracy, achieving the highest SR and lowest DtG/NRE across both 2.5D and 3D scenes. In 2.5D, NavVLM and CONVOI perform comparably due to structured sampling, whereas Spatial underperforms in all cases, highlighting the difficulty of inferring spatial scale directly from raw images. These results confirm that MVA provides essential geometric priors that enhance spatial grounding, particularly in cluttered 3D environments.

\subsection{Short-term and Long-term Navigation}
% The detailed context prompt and the corresponding response from the VLM are presented in the Appendix.
\Cref{tab:navigation} compares our method with baseline approaches across short- and long-term navigation tasks. In short-term scenarios, our approach consistently achieves the highest SR and SPL with the fewest VLM promptings. This advantage arises from the proposed Multi-modal Visual Annotation (MVA), which constrains the action space with geometric priors and adaptively refines candidate anchors, enabling more reliable decisions. Among the baselines, CONVOI performs competitively in 2.5D tasks by leveraging occupancy maps, while NavVLM obtains shortest distances to the goal but suffers from low SR due to missing yaw adjustments.

In long-term tasks, only our method and NavVLM successfully complete navigation, with our approach yielding higher SR and SPL. The longer paths arise because, when facing explored areas or infeasible decisions, our adaptive mechanism uses geometric priors to relocate the UAV for renewed reasoning.
In contrast, Spatial frequently produces infeasible commands without geometric priors, leading to early termination.

Overall, our method achieves average improvements of 25.7\% (SR) and 17.3\% (SPL) in 2.5D, and 39.3\% (SR) and 24.7\% (SPL) in 3D over the second-best method.

To further quantify each component during navigation, we conduct an ablation study on Kilburn 2.5D (\Cref{tab:ablation_kilburn}). Removing either module degrades SR and SPL. Specifically, without ADM the prompt count rises and SPL drops, confirming that validation and switching effectively prevent revisits and infeasible actions. Without MVA, SR and DtG degrade to the level of Spatial, demonstrating that geometry-informed visual annotation is essential for providing the spatial scale information that VLMs lack.
\subsection{Performance Across Different VLMs}
\Cref{tab:vlm_performance} summarizes results across four VLMs. GPT-4o achieves the best overall performance in both environments (SR = 0.67/0.57, SPL = 0.58/0.18), while Sonnet4 shows competitive SPL in Warehouse 3D (0.32). Qwen2.5 and Gemini2.5 exhibit lower SR and SPL despite stable prompt usage.

%有一些数据统计下

\subsection{Real-world Deployment}
We deployed the proposed method on a customized UAV and successfully achieved autonomous navigation to Room~407. 
During the entire navigation process, the UAV interacted with GPT-4o for five prompting rounds, resulting in a total trajectory length of 26.24\,m and a distance-to-goal (DtG) of 2.03\,m. 
The doorplate of Room~407 eventually appeared within FOV. 
As shown in \Cref{fig:realworld_demo}, the VLM selected frontier anchors (green) as navigation targets in Steps~1--3, indicating relatively low confidence in recognizing Room~407 at that stage. 
This allowed the UAV to pass around the corridor corner and effectively explore the unknown region through single-step decisions. 
In Steps~4--5, after entering the long corridor where Room~407 is located, the VLM switched to selecting target anchors (red) to approach the room.

% \subsection{Discussion}

\section{Conclusion}
\label{sec:Conclusion}
% \vspace{-1mm}
We presented a zero-shot VLM reasoning framework for adaptive UAV task-centric navigation. 
By combining multi-modal visual annotation with adaptive decision making, our method maps high-level language outputs into executable UAV actions and switches to geometric strategies when semantic cues are weak. 
The proposed method outperforms baselines in prompting efficiency, SR, and SPL, and is further validated through deployment on a real UAV.

Despite promising results, the current system reasons only at discrete stops rather than during motion. Real-world evaluation is currently limited to a single environment due to battery, safety, and flight-regulation constraints; cloud-based VLM latency is also not optimized as it depends on external API providers; extensive multi-scenario real-world trials with statistical analysis are planned as part of a journal extension. Future work will incorporate semantic priors, onboard lightweight VLMs for continuous reasoning, and scene-adaptive fine-tuning (e.g., low-rank adaptation) to further improve navigation robustness.

\bibliographystyle{IEEEtran}
\bibliography{main}
\end{document}